\documentclass[conference]{IEEEtran}
\IEEEoverridecommandlockouts

\usepackage{natbib}
\usepackage{amsmath,amssymb,amsfonts}
\usepackage{graphicx,color}
\usepackage{textcomp}
\usepackage{xcolor}
\usepackage{hyperref}
\usepackage{cleveref}
\hypersetup{hidelinks=true}
\usepackage{algorithm,algorithmicx}
\usepackage{algpseudocode}
\def\BibTeX{{\rm B\kern-.05em{\sc i\kern-.025em b}\kern-.08em
    T\kern-.1667em\lower.7ex\hbox{E}\kern-.125emX}}
\AtBeginDocument{\definecolor{tmlcncolor}{cmyk}{0.93,0.59,0.15,0.02}\definecolor{NavyBlue}{RGB}{0,86,125}}

\usepackage{amssymb}            
\usepackage{mathtools}          
\usepackage{mathrsfs}           
\usepackage{graphicx}           
\usepackage{subcaption}         
\usepackage[space]{grffile}     
\usepackage{url}                

\usepackage{amsmath}
\usepackage{mathrsfs}
\usepackage{booktabs}
\usepackage{verbatim}
\usepackage{siunitx}
\usepackage{tikz}
\usepackage{pgfplots}
\usepackage{multirow}
\usepackage{todonotes}
\usepackage{wrapfig}

\usepackage{xspace}

\newcommand{\red}[1]{\textcolor{red}{#1}}
\newcommand{\blue}[1]{\textcolor{blue}{#1}}

\makeatletter 
\newcommand{\linebreakand}{%
  \end{@IEEEauthorhalign}
  \hfill\mbox{}\par
  \mbox{}\hfill\begin{@IEEEauthorhalign}
}
\makeatother

\input{mathOperators}

\begin{document}


\title{Meta-Learning Multi-armed Bandits for Beam Tracking in 5G and 6G Networks}

\author{\centering
\IEEEauthorblockN{Alexander Mattick}\IEEEauthorblockA{Division of Positioning and Networks,\\ Fraunhofer Institute for Integrated Circuits IIS,\\ 90411 Nuremberg, Germany}\and%
\IEEEauthorblockN{George Yammine}\IEEEauthorblockA{Division of Positioning and Networks,\\ Fraunhofer Institute for Integrated Circuits IIS,\\ 90411 Nuremberg, Germany}\and%
\linebreakand 
\IEEEauthorblockN{Georgios Kontes}\IEEEauthorblockA{Division of Positioning and Networks,\\ Fraunhofer Institute for Integrated Circuits IIS,\\ 90411 Nuremberg, Germany}\and%
\IEEEauthorblockN{Setareh Maghsudi}\IEEEauthorblockA{Ruhr University Bochum,\\ 44801 Bochum, Germany} \and%
\linebreakand 
\IEEEauthorblockN{Christopher Mutschler}\IEEEauthorblockA{Division of Positioning and Networks,\\ Fraunhofer Institute for Integrated Circuits IIS,\\ 90411 Nuremberg, Germany}
}

\maketitle

\begin{abstract}
Beamforming-capable antenna arrays with many elements enable higher data rates in next generation 5G and 6G networks. In current practice, analog beamforming uses a codebook of pre-configured beams with each of them radiating towards a specific direction, and a beam management function continuously selects \textit{optimal} beams for moving user equipments (UEs). However, large codebooks and effects caused by reflections or blockages of beams make an optimal beam selection challenging. In contrast to previous work and standardization efforts that opt for supervised learning to train classifiers to predict the next best beam based on previously selected beams we formulate the problem as a partially observable Markov decision process (POMDP) and model the environment as the codebook itself. At each time step, we select a candidate beam conditioned on the belief state of the unobservable optimal beam and previously probed beams. This frames the beam selection problem as an online search procedure that locates the moving optimal beam. In contrast to previous work, our method handles new or unforeseen trajectories and changes in the physical environment, and outperforms previous work by orders of magnitude.
\end{abstract}


\section{Introduction}\label{section:introduction}

The envisioned transition to 5G and 6G technologies has started to transform the properties of established communication networks with very high transmission rates that are realized through usage of with Multiple Input Multiple Output (MIMO) antennas. However, the signal strength of the millimeter and terahertz signals used becomes too low to be reliably detectable by receivers.  Hence, at the core of this transformation also lies the capability of Base Stations (BSs) to utilize directional beams for transmission towards a User Equipment (UE) served by the network. These narrowly targeted beams allow for more efficient usage of the frequency spectrum.

Usually, a (large) library of predetermined directional beams (i.e., a \textit{ codebook}) is available, and the question is which beam needs to serve a specific UE at any point in time. The beam tracking~\citep{Giordani2018} problem is even more complicated. As UEs are also moving along unknown trajectories, the BS needs to decide (in real-time) which beam needs to be used for tracking the UE.

\begin{figure*}[t!]
    \centering
    \centerline{\includegraphics[width=\linewidth]{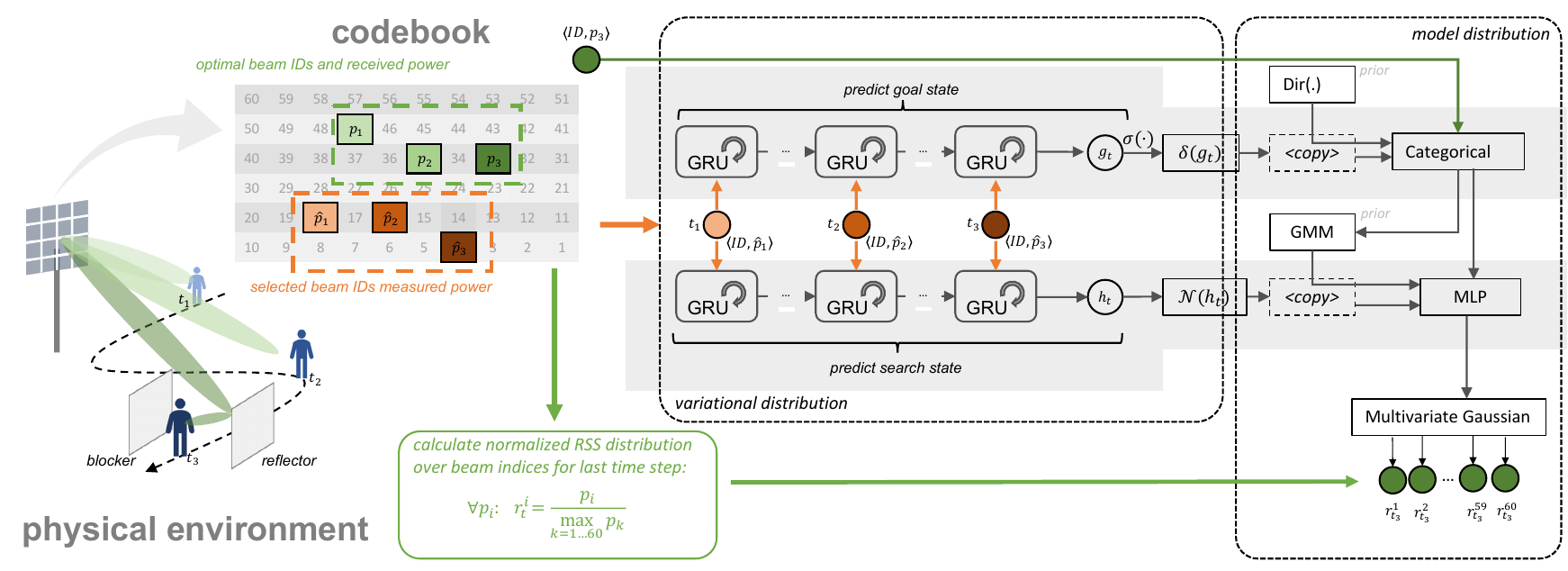}}
    \caption{Schematic view of our approach: an agent (carrying a UE) moves through the environment. The line-of-sight (LOS) between the UE and the base station might be blocked and signals might be received via reflectors. During UE tracking, we select beam IDs from the codebook (orange) while the optimal beams are shown in green (darkness of colors encode larger timestamps). The right panel illustrates the learning process based on available trajectories: the left side depicts the variational distribution, and the right side shows the model distribution. The upper half models the unobservable \textit{goal state} of the problem, while the lower half represents the \textit{exploration state}.}
    \label{fig:intro}
\end{figure*}

The core challenge of beam management is that the location of the UE is usually unknown. Even if we would first try to locate the UE (which itself is not trivial as positioning requires a coordinated procedure involving several BSs and the network) we still do not know the surrounding environment with all its material properties and their effect on radio waves propagation -- this would be necessary for e.g. a digital twin solution where beams are selected based on a path tracing from the BS to the UE.
In fact, even if we had such an accurate 3D scan of the environment (with absorptive/reflective properties), we cannot guarantee that this scan is going to be up-to-date with recent changes in the environment, e.g., trees with/without leaves, parking cars, etc.
Beyond that, beam management has strong real-time constraints: If one assumes a non-stationary UE, then the entire beam-selection problem has to be solved before the UE has moved from its current position. 
This has lead to a reliance on brute-force beamsweeps to select optimal beams, which is both energy-intensive and not scalable to large codebooks.

We propose a novel method relying on Reinforcement Learning (RL) that aims to select the correct beam solely based on the received signal strength (RSS) feedback rather than using the ground truth position or 3D models.
We utilize a temporal prediction system that exploits the locality of the UE's movement, without needing explicit or implicit location estimation.
To do this, we employ the framework of the ``Restless Multi-Armed Bandit'' (RMAB) to facilitate online search using a pre-trained neural network.
As our method only uses a small stochastic neural network as a predictor, our beam selections are fast, without needing any raytracing or brute-force search.
We show that our approach is robust against transfer to unknown trajectories and unseen environments and that we outperform existing RL approaches on the order of magnitudes in the number of beams to be probed to achieve high reference signal received power (RSRP).

The paper is structured as follows. We discuss related work in Sec.~\ref{section:related work}. Sec.~\ref{section:background} provides background information on POMDPs, the  RMAB framework, and beamforming. 
Sec.~\ref{section:meta_learning-mab} formalizes the problem, 
describes our model, and discusses its instantiation.  
Secs.~\ref{section:experimental-setup} and~\ref{section:evaluation} describe the experimental setup and results. Sec.~\ref{section:conclusion} concludes.

\section{Related Work}
\label{section:related work}

There has been significant work trying to improve upon naive beam-sweeps for beam selection.
\citet{SonyLSTM} uses an LSTM and supervised learning to regress on the correct beam-id.
\citep{Maggi2024ToRO} provide a great overview on when to use RL and when to use other (e.g. supervised or stochastic programming) techniques in radio communications.
\citet{Maggi2023TrackingTB} utilize a sophisticated Bayesian optimization framework to select beams in a Bayes-optimal sense. They first sample $M$ beams and then build a Gaussian Process Model to give a stochastic estimate over the codebook entries. The selection is now performed according to a parallelized variant of the expected-improvement acquisition function.
\citet{ThompsonBeamMan} use traditional Thompson sampling (i.e., time-independent search) to locate the optimal beam. This method is quite similar to ours, but relies on the assumption that radio signals change slowly as no dynamics (e.g., UE movement) are taken into account during beam selection. Our framework is designed for the opposite case: We assume that a UE might constantly move and therefore our search has to constantly readjust its own belief about beam quality.

RMABs were originally proposed by \citet{restlessBandits} for optimizing resource allocation using classical Lagrange-multiplier and Gittins-index methods~\citep{gittins-index} under the assumption that the dynamics are fully known.
Recent approaches study unknown dynamics, e.g.,~\citet{Liu2010LearningIA}
analyze RMABs using iterative ``exploration'' and ``exploitation'' phases.
However, theoretical guarantees can generally only be given under strong assumptions on system parameters.
Another recent area of research distinguishes between exogenous and endogenous systems~\citep{Gafni2022RestlessMB,Liu2010LearningIA}.
\citet{Meshram2017AHM} consider a POMDP scenario in which the unknown stochastic process is estimated using a Hidden Markov Model (HMM) under binary rewards.
\citet{Meshram2016OptimalRT} analyze the behavior of Thompson sampling~\citep{Thompson}, in the restless single-arm setting with binary reward.
The closest work to ours is probably \citet{Jung2019RegretBF} and \citet{Jung2019ThompsonSI} who use Thompson sampling for RMABs and a similar deploy-then-update routine.
However, in addition to their use of a deterministic policy (while we use a stochastic one) just like most prior work they assume 
the rewards are binary.

From an RL point of view, several studies have explored the problem of beam selection in millimeter-wave (mmWave) communication systems. One notable work is proposed by \citet{Raj2020DeepRL} who use Deep Deterministic Policy Gradient (DDPG) to address the blind beam selection problem, i.e., to replace traditional beam sweeping techniques. Another approach investigates the optimality and robustness of both the Deep Q-Network (DQN) and DDPG to the beam selection problem~\citep{Lee2020DeepRL}.
A closely related approach by \citet{sim2018online} uses online contextual bandits and fast machine learning (FML) for contextual multi-armed bandits. However, their approach makes significantly more assumptions on the environment (i.e., straight bi-directional road where cars send direction indicators, both being used as a static context for the bandit) and assumes vastly smaller codebooks of which 1/4th are allowed to send at once.

From the point of view of Beam selection, most systems still rely on beam sweeps across a finite codebook of beam directions. 
Practical deployments tend to rely on codebooks as the directions can be aligned well to real world hardware limitations, e.g. quantized phases.
While slow, exhaustive beam sweeps produce good results after search (see also \citep{BeamformingProtocol}).
One can further refine this by hierarchical sampling where one descends from coarse, wide beams toward narrower beams across multiple measurements (see for an overview \citep{MultiscaleSelection}).
The downside of these methods is the lack of speed as even in the hierarchical setting many low-performing beams have to be tried before a good ones are found. Further, in dynamic scenarios the optimal beam might have shifted by the time the UE is located.
In our problem we assume a fast moving UE (or, analogously narrow beams), meaning that only one (or a small number) beam can be trialed in each iteration before the UE position has changed.
This reduces exhaustive search to essentially a random search, since exhaustive exploration is stateless between UE position changes.

More involved approaches utilize optimization algorithms that may be coupled to the HW properties of a specific basestation.
For instance, \citet{app122312282} uses manifold optimization to build a hybrid analog-digital beamforming system.
While these methods are undoubtedly interesting, the analysis of the hardware-software co-design of real world basestations is out of scope for our work.

While these prior works have made significant contributions to the field, it is important to note that deterministic methods may be suboptimal in scenarios where predictions involve uncertainty, particularly when considering unknown receiver positions. In mmWave communication systems, the probabilistic nature of predictions plays a crucial role due to factors such as channel fading, noise, and environmental changes. Therefore, it is imperative to account for the inherent uncertainty and adaptability requirements when addressing the beam selection problem.

\section{Problem Description}
The beam selection problem is a necessary sub-task for beamforming antenna arrays:
Given a fixed codebook $\mathcal{A}$ of $N$ possible beam directions to choose from, the beam selection problem involves finding the beam $a\in\mathcal{A}$ that maximizes the Received Signal Strength (RSS) to a UE with a minimal amount of measurements of codebook entries.
Ideally, one would have access to the UEs location and moving direction, as well as a 3D model of the environment including material properties to facilitate e.g., raytracing based solutions, but this is usually not possible:
In 3GPP, the UEs location and movement direction is not available for beam selection due to privacy concerns, and even if those were available gathering a 3D model with material properties is intractable.

For this reason, we assume a black-box structure, where we have no information about the environment, reflections, or even the azimuth and elevation of each beam in the codebook.
To solve this problem, we search for an optimal beam by choosing $k=1$ beams from the codebook and update our choice based on the measured RSS.
Since UEs usually do not have a fixed position, we also predict a correction factor $g$ which is used by the beam-choice algorithm to correct for the movement of the UE.
This way, the beam selection problem decomposes into a state-estimation problem for the correction factor $g$, and a decision problem for finding the optimal beam.
%
This optimization problem is the typical scenario for Reinforcement Learning (RL), where one aims to maximize the long-term discounted cumulative reward (RSS) just by interacting with the environment.

\section{Background}
\label{section:background}
\subsection{Reinforcement Learning}\label{sec:RLBackground}

Reinforcement Learning (RL) describes a field of study that attempts to solve sequential Markov Decision Processes (MDPs).
An MDP consists of a set of states $s\in\mathcal{S}$, actions $a\in \mathcal{A}$, reward $R:\mathcal{S}\times\mathcal{A}\to\mathbb{R}$, a transition distribution (typically unknown) $T:\mathcal{S}\times\mathcal{A}\to\mathcal{S}$, and a discount factor $\gamma$. We train a decision making policy $\pi:\mathcal{S}\to\mathcal{A}$ that selects actions in states. Our objective is to find the optimal policy $\pi^*$, which maximizes the expected discounted sum of future rewards:
\begin{equation*}\label{eq:MDP}
    \pi^* =\operatorname{argmax}_\pi \mathbb{E}\left[\sum^{t_{\text{max}}}_{t=0} \gamma^t R(s_t,a_t)\pi(a_t|s_t)T(s_{t+1}|s_t,a_t)\right],
\end{equation*}
where $0\leq\gamma<1$ can be configured to trade immediate over future rewards and guarantees boundedness of the expectation~\citep{Sutton1998}. 
In our case, we assume a POMDP setting, which further differentiates between observations $O_{1:t}$ and states $s_{1:t}$. 
``Observations'' are measurements we directly get from the environment (e.g. RSS values), while ``states'' include unobserved information, such as the UE's ground truth position.

In this work we follow a standard POMDP formulation:
A POMDP is defined as a 7-tuple $(\mathcal{S},\mathcal{A},T,R,\Omega, \mathcal{O}, \gamma)$.
The definitions are largely the same as in the ordinary MDP with the extension of an explicit observation set
$\Omega$ and conditional observation probabilities conditioned on the current (hidden) state  $\mathcal{O}:\mathcal{S}\times \mathcal{A}\times \Omega\to\mathbb{R}$.
In other words, $\mathcal{O}$ is the probability of observing $O\in\Omega$ in state $S\in\mathcal{S}$ after executing action $A$, and $\gamma\in [0,1)$ is the reward discount factor.

POMDPs reduce to ordinary MDPs in the case of $\mathcal{S}=\Omega$, but if $\mathcal{S}\neq \Omega$, the agent only observes partial information of the true state.
This makes POMDPs significantly more challenging to solve as the agent also has to account for the uncertainty induced by the necessary state estimation. Let $b:S \to \mathbb{R}$ be the \emph{belief state} over a state $s$. The agent maintains the belief state by updating it with received observations and actions:
\begin{equation}
    b(s') \propto \mathcal{O}(o \mid s',a)\sum_{s\in S} T(s' \mid s,a)b(s),
\end{equation}
as derived in \citep{KAELBLING199899}.

This problem generally becomes difficult in practice due to the fact that the exact observation probabilities $\mathcal{O}(o \mid s',a)$ and transition probabilities $T(s' \mid s,a)$ are unknown.
In the case of beam selection, the unknown part of the state is the position/change of position of the UE.

\subsection{Multi-Armed Bandits (MABs)}\label{sec:MAB}

The multi-armed bandit (MAB) problem is formally equivalent to a one-state Markov decision process. Here, we assume a stationary environment with respect to an agent's actions: no matter which action $a\in\mathcal{A}$ the agent takes, it will not change the state.
In the MAB setting, the primary objective is to estimate the action (referred to as an ``arm'') that maximizes the expected utility/reward.
However, it is also crucial to balance exploration and exploitation, meaning that the agent must actively explore different actions to optimize its long-term performance.
The MAB model is built to learn the optimal tradeoff between discovering new high performing actions, while also maintaining high cumulative rewards.

If one assumes full access to the payoff distributions, one can solve MABs directly using an index policy known as the Gittins-index~\citep{gittins-index}.
However, since in practice these are not known, one can use highly efficient approximations, such as Thompson-sampling (TS)~\citep{Thompson}, which updates the expected utility of each arm using Bayes' rule.
The high efficiency makes MABs very popular in online decision-making scenarios such as recommender systems~\citep{Zhou2017LargeScaleBA}, black-box optimization~\citep{dimmery2019shrinkage}, anomaly detection~\citep{Ding2019InteractiveAD}, healthcare~\citep{pmlr-v85-durand18a}, and machine learning~\citep{Gagliolo2008AlgorithmSA}.

The limitation of MABs compared to ``normal'' Reinforcement Learning is that one has to assume stationarity in the payoff distribution. Stationarity here refers to the distribution behind each arm not changing during sampling, which makes modeling dynamic problems (such as moving UEs) impossible.
Fortunately, many extensions to the MAB problem exist that relax this assumption, allowing one to use variants of the MAB algorithms on dynamic problems.
One of the most unrestricted these realizations is the Restless MAB.

\subsection{Restless MABs (RMABs)}

We utilize an extension of the MAB framework, sometimes referred to as a ``restless'' MAB. In this extension, we assume that there exists an unobservable stochastic transition system $P(s_{t+1} \mid s_t)$ that rotates through hidden states $s\in\mathcal{S}$ and that changes the payoff distribution behind each arm.
More precisely, one assumes the existence of Markov transitions $P_a,P_{\bar a}$ that change all arm distributions depending on whether arm $a$ was selected ($P_a$) or not selected ($P_{\bar a}$).
Note that RMABs are only partially observable as we only ``see'' the outcome of arms we actually sample.

Solving RMABs is computationally intractable, as optimal policies are NP-hard to compute even in the finite horizon setting~\citep{Papadimitriou1994TheCO}.
One possible heuristic is to study a time-averaged version of RMABs which yields the Whittle-index~\citep{restlessBandits}.
Unfortunately, the Whittle-index is itself generally only a heuristic\footnote{For instance, one has to guarantee indexability, which corresponds to the set of unplayed arms growing monotonically as the underlying lagrangian increases.}. 
Further, because the index is computed offline and fixed, any model mismatch or estimation error persists - there is no inherent mechanism to adjust the index in light of new observations.
This implies that finding an index policy is hard to frame as a machine learning problem: Exploration vs Exploitation of different indices is not trivial to include in index policies.

Recent advancements in learning RMABs suggest that Thompson Sampling~\citep{ThompsonSelfCorrecting}, which is an asymptotically optimal Bayesian technique for solving ordinary MABs, can be trivially adapted towards a close to optimal RMAB algorithm~\citep{Jung2019RegretBF} in the episodic setting.
The Bayesian nature of TS and the episodic setting makes this a prime candidate for the introduction of machine learning into RMAB solving.

\section{Beam Selection as RMAB}

In this work, we model the beam tracking problem as an RMAB by considering the set of arms $\mathcal{A}$ to be the codebook of beams (see, for instance, the upper left part of Fig.~\ref{fig:intro}: the codebook consists of beamforming vectors (or precoding vectors), each of them corresponding to a specific beam direction or pattern), $\mathcal{S}$ to be the description of the location of the UE and the RSS value of all beams in the codebook, while the arm transitions $P_a,P_{\bar a}$ are (implicitly) given by the movement of the UE. 
Note that for beam selection, we can assume that $P_a=P_{\bar a}$ since a UE's movement can be assumed to be independent of which beam was selected.
This means that while we still have a state-evolution in all arms (even the ones that have not been tried), the state change is independent from the arm selection.
The major challenge in treating beam tracking as an RL problem is that we only observe the part of the state (beam RSS) we end up measuring while the rest of the state remains unknown: Beam selection is a POMDP.

Since solving such an RMAB is computationally intractable, we treat the RMAB as an episodic task similar to \citep{Jung2019RegretBF} which allows us to use a machine learning approach to solve the RMAB:
Specifically, we split the RMAB inference into two parts.
First, we collect $N$ episodes by running TS in simulation and train a recurrent neural network to predict TS’s action distribution by updating the parameters $\theta$ of a stochastic neural network.
Second, during deployment, the resulting network is now capable of approximating the optimal beam distribution from just the sequence of previously tried (beam-id, rss) values, \emph{without needing any further training} (online inference phase).

Since we train a suitable network during the offline training phase, the online inference phase is as fast as a single forward pass through the neural network.

\section{Meta-Learning Multi-armed Bandits}
\label{section:meta_learning-mab}

For our beam-selection problem, we consider an $N$-armed bandit, where $N$ is the size of our codebook, and ``pulling'' an arm $a_i$ means measuring the received power $r_i$ at arm $a_i$. 
Unfortunately, naive application of existing approaches runs into the issue of distribution shift: As the UE moves through the environment the hidden state $s_t$ determining the arm-payoff keeps changing.
Accounting for this change in arm-payoff necessitates using a more complex arm distribution model, which is intractable to solve online using \eg MCMC.
The following sections will cover how we solve the distribution shift (Section~\ref{sec:distributionshift}) and the inference complexity issues (Section~\ref{sec:meta-learning}).

\subsection{Distribution shift in the underlying reward process}\label{sec:distributionshift}

In contrast to ordinary MABs where the arm distribution is assumed to be \emph{stationary}, the beam quality distribution shifts as the unobservable UE moves.
We solve this with an estimator $P^g(\ve{g}_{t+1} \mid \ve{o}_{1:t})$ that attempts to predict the optimal beam ID given all previous observation up until the current time step. This tries to account for the distribution shift in the Multi-Armed Bandit distribution induced by the movement of the UE. We approximate a second distribution $P^s(\ve{r_{t+1}}(a) \mid \ve{o}_{1:t}, \ve{g}_{t+1})$ that, intuitively, acts as the actual MAB process that aims at solving the exploration-exploitation trade-off, i.e., accurately predicting the expected signal strength $\ve{r_{t+1}}(a)$ of beam ID $a$, given all previous observations and the prediction of the optimal beam ID. The resulting RMAB model is: 
\begin{align}
\begin{split}
    P(\ve{r_{t+1}}&(a), \ve{g}_{t+1} \mid  \ve{o}_{1:t}) =\\
    &\underbrace{P^s(\ve{r_{t+1}}(a) \mid \ve{o}_{1:t}, \ve{g}_{t+1})}_{\text{bandit ``search'' head}} \underbrace{P^g(\ve{g}_{t+1} \mid \ve{o}_{1:t})}_{\text{prediction ``heatmap''}}\; ,\label{eq:factorizedModel}
\end{split}
\end{align}
where $\ve{o}_{1:t}$ is the set of observed values up to timestep $t$, $\ve{g}_{t+1}$ is the estimated optimal beam id at time step $t+1$, and $\ve{r}_{t+1}(a)$ is the reward distribution at time step $t+1$ when choosing action $a\in\mathcal{A}$.

The main purpose of $P^g$ is to account for the distribution shift induced by a moving receiver, with the goal of correcting the $P^s$ distribution such that the sampling process becomes a static distribution estimation problem from the point-of-view of the search head $P^s$.

\subsection{Meta-Learning ahead of time}\label{sec:meta-learning}

The second problem is how to make the distribution estimation tractable.
Normally, Bayesian estimation through methods such as Markov-chain Monte-Carlo (MCMC) \citep{tierneyMarkovChainsExploring1994} or Stein's variational descent \citep{liuSteinVariationalGradient2019}, become intractable to do online due to the large number of iterations they require, especially when using uninformative distributions as starting points.
At worst, as the UE is constantly moving it might have already moved away from the selected beam by the time inference is complete.
Hence, we propose \emph{meta-learning} a distribution ahead of time, such that updating the estimate with further observations $\ve{o}_{t+1}$ is only as difficult as executing a stochastic function.

For this reason, we propose using a recent approach known as stochastic variational inference (SVI)~\citep{SVI} to generate the distributions.
SVI approximates the intractable posterior of the latent variable $z$ given the observed values $x$ (from some dataset),
using a different, parameterized distribution $Q_\phi$, which is obtained by minimizing the KL divergence between $P$ and $Q$:
\begin{equation}
   \min_{\bar{\phi}, \bar{\theta}} D_\mathsf{KL}(Q_{\bar{\phi}}(z) \parallel P_{\bar{\theta}}(z \mid x)) \;. \label{eq:KL_div_var_dist}
\end{equation}
Note that the variational distribution $Q$ does not depend on the elements of the dataset $x$, making it trivial to sample from, without requiring the intractable distribution $P(x)$. Instead, the information on the dataset $x$ is amortized into the variational parameters $\phi$ during training, which can be achieved efficiently by minimizing the KL-divergence or equivalently maximizing the Evidence Lower Bound (ELBO):
\begin{align}
\begin{split}
    \operatorname{ELBO}& = \mathbb{E}_{Q(z|o_{1:t})}[\log (P^s_\phi(r_{t+1}(a)|z,g_{t+1}))+\\
    \log(&P^g_\phi(g_{t+1}|z))+ \log(P(z))- \log Q_\theta(z|o_{1:t})],
\end{split}\label{eq:ELBO}
\end{align}
where $P(z)$ is an arbitrary prior over the latent variables $z$.
We use this SVI pipeline to train a policy $\pi$ which is then \emph{fixed during deployment}, effectively offloading the complexity of updating the posterior away from the inference time onto training time.

\subsection{Online Inference}

For the case of online inference, we need the previous observations $\ve{o}_{1:t}$ as input since those are the observations that we will have when exploring the environment online.
The KL divergence between variational and model distribution from Eq.~\eqref{eq:KL_div_var_dist} is only valid if we assume that only the data provided in the dataset is relevant to the prediction (as only those get amortized into the parameters).
In the case of online prediction of some datapoint $\ve{o}_{1:t}$ we also have to condition our distribution on the observations.
SVI allows us to amortize the dataset $x$ into the trainable parameter.
In our case, we can optimize this distribution by simply conditioning $P$ on the current set of observations $\ve{o}_{1:t}$.
An observation $\ve{o}$ consists of both the received power and the measured beam, giving us $\ve{o} = (\text{RSS}, \text{beam-id})$.

The objective containing the online conditioning on $\ve{o}_{1:t}$ is described by
\begin{align}
\label{eq:online}
     \nonumber \min_{\bar{\phi}, \bar{\theta}} D_\mathsf{KL}(&Q_{\bar{\phi}}(z \mid \ve{o}_{1:t},\ve{r}_{t+1}(a)) \parallel \\ 
     &P_{ \bar{\theta}}(z\mid \ve{o}_{1:t},x,\ve{r}_{t+1}(a),\ve{g}_{1:t})) \;.
\end{align}
In words: We align the distribution $Q_{\bar\phi}$ that does not know $x$ to $P_{\bar\theta}$, which does know $x$.
This naturally forces the information in the dataset $x$ into the parameters $\bar\phi$ and $\bar\theta$.

Since we have a goal state $g_{t+1}$ and an arm sampling state $r_{t+1}(a)$, we also partition the latent along the same line $z=(z_g,z_s)$.
The latent variable random $z$ learned using this objective can now be used inside the distribution to 
\begin{align*}
P(\ve{r}_{t+1}(a),\ve{g_{t+1}} \mid \ve{o}_{1:t}, x) \propto& P^s_{\bar\theta}(\ve{r}_{t+1}(a)\mid z_s, \ve{g}_{1:t+1})\\ 
&P^g_{\bar\theta}(\ve{g_{t+1}} \mid z_g)Q_{\bar\phi}(z_g,z_s|\ve{o}_{1:t}),
\end{align*}
where $\ve{g}_{t+1}$ is the ``goal'' beam, i.e. the best-beam estimate used to correct for UE movement, $z$ is the latent parameter, and $\ve{r_{t+1}}(a)$ is the distribution of returns for arm $a$.
Note that our model is factorized into the two components ``search'' and ``goal'', as shown in Eq.~\ref{eq:factorizedModel}.

The information from the dataset $x$ here is amortized into the static weights $\theta$ and $\phi$.
The model can now be trained by executing the current distribution in the environment using Thompson sampling~\citep{Thompson}, storing the gathered datapoints in our dataset, and training the next distribution on the extended dataset~(see Fig.~\ref{fig:intro}).

During inference, we treat the parameters $\theta$ and $\phi$ as static, and simply choose the maximal arm from a vector of samples from the learned $P^s_\theta(\ve{r_{t+1}}(a)|z,\ve{o}_{1:t})P^g_\phi(z \mid \ve{o}_{1:t})$ distribution, see Thompson sampling~\citep{Thompson}.

\subsection{Practical Considerations}
\label{section:MLpractical}

We approximate the beam selection process hierarchically and chose to model the prediction heatmap $P^g(\ve{g}_{t+1} \mid \ve{o}_{1:t})$ as a categorical distribution with an uninformative Dirichlet prior with all $\alpha_i=1$.
The bandit search head $P^s(\ve{r}_{t+1}(a) \mid  \ve{o}_{1:t}, \ve{g}_{t+1})$ is implemented as a three layer MLP with skip-connections that receives both the categorical prediction $g_{t+1}$ and the observations $\ve{o}_{1:t}$ in the form of a hidden state $h_t$ with a Gaussian mixture model prior (see Fig.~\ref{fig:intro}).
To predict the posterior $z$ consisting of the goal-latent $g_t$ and the search-latent $h_t$, we utilize two gated recurrent units (GRU)~\citep{GRU} that receive the sequence of previous observations as inputs. 
For increased stability, we normalize the received power measurements ``per frame'' meaning we divide the RSS by the maximally receivable power at that timestep
\begin{equation}
    r_t(i) = \frac{p_i}{\max_{k=1\dots N}p_k} \;, 
\end{equation}
where $p_i$ is the raw measured RSS for beam ID $i$ and the denominator refers to the RSS of the best beam ID (i.e., the highest RSS that is achievable).
The objective of this is to prevent the model from being able to completely ignore learning the dynamics of states where all codebook entries have poor RSS values.
Note that the model remains trainable if we remove the normalization, but the convergence is slower. 
We only normalize the reward but not the observation: During deployment we only measure individual beams, which means we cannot normalize based on the best beam. To not introduce a domain shift between training and inference, we keep the observations unnormalized.
\begin{algorithm}[b!]
    \begin{algorithmic}
        \Function{act}{$P_\phi$, $Q_\theta$, $o_{1:t}$}
            \State  $z\sim Q(z|o_{1:t})$ 
            \State  $g_{t+1}\sim P^g(g_{t+1}|z)$ 
            \State  $r_{t+1}(a) \sim P^s(r_{t+1}(a)|z,g_{t+1})$ 
            \State  select beam $\operatorname{argmax}_a r_{t+1}(a)$
            \State  observe RSS for $o_{t+1} = (a,RSS)$ 
            \State  during training: also observe $g_{t+1}$
        \EndFunction\\
        \Function{train}{$P_\phi$, $Q_\theta$, $o_{1:T}$, $r_{1:T}(a)$, $g_{1:T}$}
        \ForAll{$t\in\{1,\dots,T\}$}
            \State  Estimate ELBO using Eq~\ref{eq:ELBO}
            \State  update $\phi,\theta$ using Adam.
        \EndFor
        \EndFunction
    \end{algorithmic}
    \caption{Pseudocode of our method: During acting, we sample according to the currently learned distribution and record all actions, observations, and goal states in a buffer. Batches from the buffer $o^N_{1:T}$, $r^N_{1:T}(a)$, $g^N_{1:T}$ are later used during training to optimize the parameters $\phi,\theta$.}\label{alg:rl_framework}
\end{algorithm}

To further increase stability we use automatic Rao-Blackwellization in the stochastic programming library pyro (see e.g. \cite{Schulman2015GradientEU} and \cite{Rao-blackwell-theorem}) and a batchsize of 64 full trajectories consisting of 200 steps each.
For exploration, we utilize a simulated environment with 12 parallel agents and a total exploration budget of $500,000$ steps in our simulated environment.

To account for the fact that our exploration dataset grows as we continue exploring, we scale the number of training steps after each round of exploration to 
$$steps=\left\lfloor C\sqrt{\frac{|\mathcal{D}|}{\text{batchsize}}} \right\rfloor,$$ 
where $|\mathcal{D}|$ is the current dataset size.
We set $C=3$ to roughly match the expected number of training iterations to the number of environment rollouts in each training episode.
This is not an extensively tuned parameter and is only used to align the number of updates to the dataset size and therefore reduce the total amount of compute necessary.
We use the ADAM optimizer~\citep{Kingma2014AdamAM} with a learning rate of $3\times 10^{-4}$. Algorithm~\ref{alg:rl_framework} shows the pseudo code of our algorithm.

\section{Experimental Setup}
\label{section:experimental-setup}

\begin{figure}[t!]
    \centering
    \includegraphics[width=\linewidth]{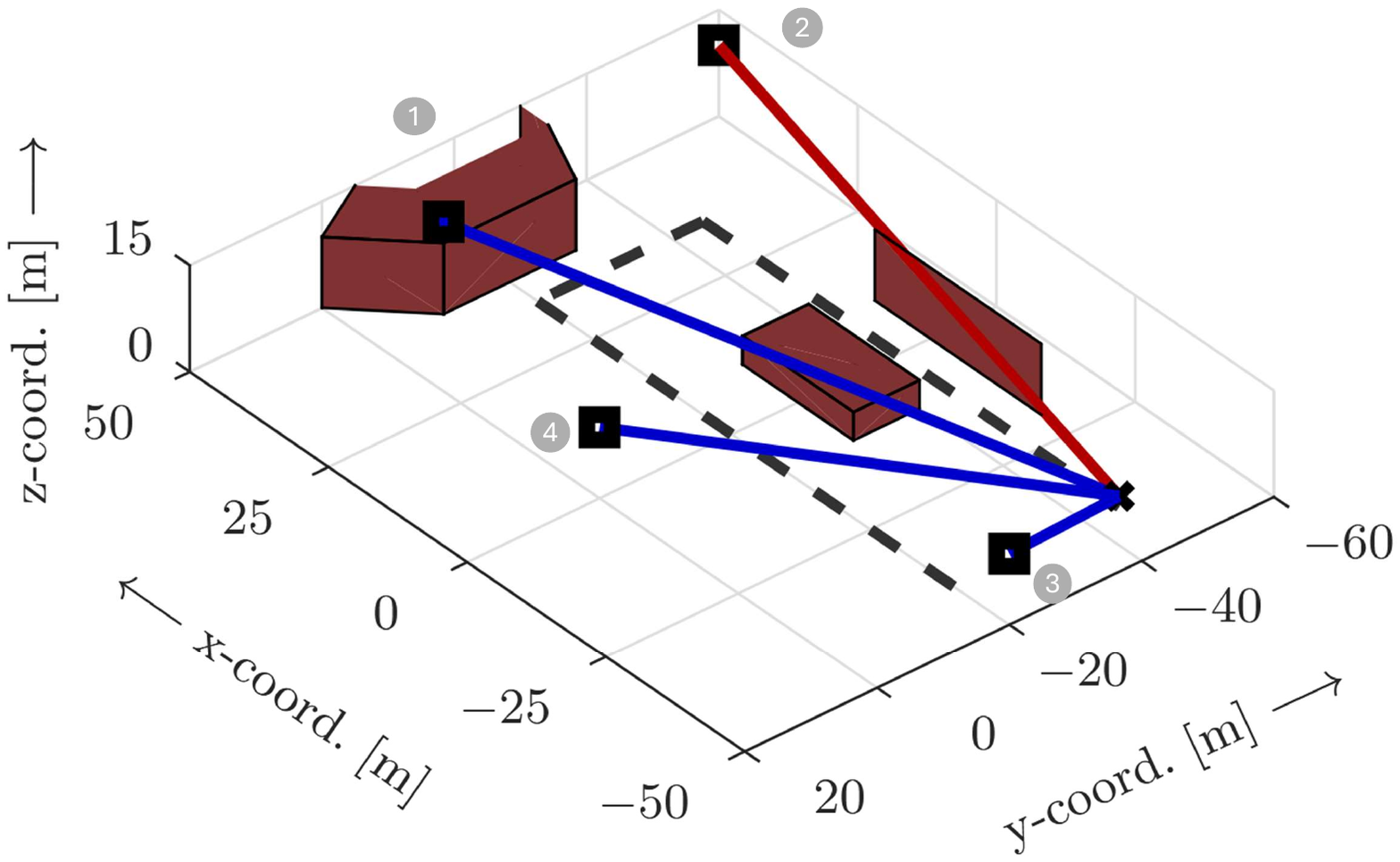}
    \caption{Visualization of the simulation environment: The UE (black cross) moves around the environment following the dashed path. At each step, the link between the BSs (black square) and the UE is checked for obstructions. Here, if a link is obstructed it is plotted in red and blue otherwise.}
    \label{fig:sim_env}
\end{figure}

\textbf{Geometries.} We consider three slightly different environments for our benchmark: First an environment with a single basestation and flexible real-world configurations (e.g. with/without certain signal blockers and reflectors). This is aimed to show robustness against real world variation.
Second, we consider a setting with 4 basestations and different codebook sizes to investigate scaling across different numbers of beams and basestation positions.
Finally, in order to also investigate moving the UE at different speeds through the environment we instantiate an environment with different total \textit{speeds}.

\textbf{Simulation environment and geometry.} We designed and implemented a simulator to generate the required radio channels and to obtain the input RSS values. We set up a simulation environment in \textsc{Matlab} with a geometry that is representative of a typical urban scenario. We place four base stations in the environment and let their panels point into a specific directions, see Fig.~\ref{fig:sim_env}.

To model complex environments, we add objects that act as blockers and attenuate the received signal, or as reflectors. They are defined using different shapes, sizes and locations. When acting as blockers, signals are attenuated (with a varying amount between experiments, imitating different materials) when passing through.  
When acting as reflectors, the signal power drops to account for reflection losses. We apply such parameters based on experimental data~\citep{attenuation}.

The radio channels are generated using QuaDRiGa~\citep{quadriga1}. For the system model, we chose a carrier frequency of 28\,\si{GHz}, a bandwidth of 100\,\si{MHz}, OFDM modulation, and 64 active sub-carriers. The base stations are modeled with uniform rectangular array antenna panels with half-wavelength-separated elements. For beamforming, discrete analog DFT-based~\citep{Tse05} codebooks of predefined azimuth and elevation pointing angles are designed given a panel design similar to that from Fig.~\ref{fig:intro}.
For simplicity, the UE is equipped with an omni-directional antenna (i.e., no beamforming).

\textbf{UE trajectories and movements.} Using QuaDRiGa, we define the environment layout and place the 4 base stations and blockers. We then randomly defined a trajectory for the UE using 4 anchor points defining a 
track around the middle blocker when present, otherwise similar tracks are used when the blocker is not present in the environment. We project a ray from each BS to the UE for each spatial sampling point on this track and count the intersections with blockers and attenuate the signal accordingly. Once this info is obtained, we generate the frequency-domain channel coefficients including the environmental attenuation. Unless stated otherwise, we use 70\% of the generated trajectories for training 15\% for testing and 15\% for validation.

\textbf{Evaluation metric.} For our evaluation, we assume a system without noise and only line-of-sight (LOS) or obstructed LOS paths and no multipath signals. As a measure, we define the RSS indicator as the received power calculated in the frequency domain. Even though we do not consider more sophisticated network topologies in this work, in order to clearly compare the performance of our approach with the available baseline solutions, the task at hand remains a challenge whose complexity stems from the availability of multiple BSs with large codebooks.

\textbf{Computing infrastructure.} For all experiments we used a machine equipped with 32GB memory, an Nvidia RTX3090 GPU and an AMD Ryzen 5800X3D CPU.
We did not experience any hardware bottlenecks as we were mainly constrained by the inherent sequentiallity of our model and SVI itself.

\section{Evaluation}
\label{section:evaluation}



We investigate the robustness of the selection policies learned by our method by evaluating across unseen trajectories and environments (\Cref{sec:generalization_move_env}), codebook sizes and basestation positions (\Cref{sec:Impact_size_env}), movement speeds of UEs (\Cref{sec:movementspeedBench}),   and close this with an ablation of the size of the correction model (\Cref{sec:GoalModelAblation}).

Since we assume a small number of beam evaluations for every timestep, we do not directly compare against exhaustive beam sweeps since they would either be essentially random (one beam evaluation per timestep while the UE is moving) or would break the beam evaluation budget by an order of magnitude or more\footnote{We do report the performance of random selection and report the achieved RSS as a percentage of the optimal (exhaustive sweep) beam}.
Instead, we demonstrate the performance of our MAB-based approach by comparing it against state-of-the-art solutions from the field of communications \citep{Maggi2023TrackingTB} and Reinforcement Learning~\citep{Schulman2017ProximalPO} (\Cref{sec:SOTA_comparison}).
Additionally, we offer an uncertainty estimation experiment in \Cref{app:uncertaintyOracle}, which showcases how our method can dynamically delegate to e.g. full beam sweeps if its uncertainty is high.



\subsection{Robustness and Generalization across Trajectories and Environments}\label{sec:generalization_move_env}
\begin{table}[t!]
    \centering
    \caption{Test optimal beam classification of training and validating on different dataset combinations.
    This shows the $\text{RSS}/\text{top-1}/\text{top-5}$ accuracy of classifying the optimal beam for unknown trajectories.}
    \resizebox{\linewidth}{!}{
    \begin{tabular}{ccccc}
        \toprule
         &\multicolumn{3}{c}{tested on $60$ beams}\\\midrule
         \multirow{4}{*}{\rotatebox[origin=c]{90}{trained on}}&& no blocker & blocker & blocker ref \\ 
        &no blocker    &  176.43 / 70\% / 91\%    &  159.06 / 51\% / 73\% &  160.99 / 55\% / 78\%   \\ 
        &blocker       &  143.94 / 52\% / 75\%    &  163.54 / 53\% / 76\% &  156.25 / 52\% / 75\%   \\ 
        &blocker ref   &  120.18 / 36\% / 64\%    &  132.74 / 35\% / 62\% &  141.88 / 40\% / 70\%   \\ 
        &random        &  \ 18.12 / 2\% / 8\%    &  \ 37.76 / 2\% / 8\% &  \ 31.78 / 2\% / 8\%    \\
        \bottomrule
    \end{tabular}
    }
    \label{tab:test_accuracy}
\end{table}

We evaluate the robustness of our approach to changes in the environment. To do so, we consider only BS2 (see Figure~\ref{fig:sim_env}) and use a codebook consisting of 60 beams while varying the \emph{physical} topology of the environment. Specifically, we instantiate three versions of the environment from Fig.~\ref{fig:sim_env}: (1) without the central cuboid blocker (``no blocker''), (2) with the blocker (``blocker'', 
), and (3) with both the blocker and reflections (``blocker ref''). BS2 is placed 10\,m above the ground, pointing towards the central ``blocker''.
We train a control policy for BS2 on 3 different trajectory types, each with 350 instances for 200 timesteps each, and record all 60 codebook power indicators. 
For testing we do the same but use \emph{unseen} trajectory types.

Table~\ref{tab:test_accuracy} shows the results. We report the normalized achieved RSS and the top-1 and top-5 accuracy (i.e., if our method chooses the best or one of the top-5 best beams). As we normalize the reward per timestep, the total achievable reward is 200 (number of steps in each trajectory). Noticeable at first glance, our method performs overall well as it collects the vast majority of the achievable reward. Note that the theoretical limit of 200 is only achievable if the agent may predict the movement of the UE perfectly.
The ``random'' results correspond to a naive beam-sweeping procedure that takes 60 time steps to probe all the beams. From these results we can conclude that our method is robust against modest to large changes in the physical environment with our model only negligibly sacrificing reward.
One particularly interesting artifact is the better performance on the ``blocker ref'' environment that we obtain training without the blocker instead of training on ``blocker ref'' directly. We believe this to be due to the higher complexity of the blocker ref environment yielding slower convergence and a generally overly conservative policy.

Beyond classical robustness to changes in the domain, we are also able to quantify our method's uncertainty due to the usage of Bayesian inference.
This can be quite useful to e.g., use classical methods once the agent hits consistent high uncertainty or to trigger additional training.
We accomplish this by predicting the chance of error directly from the internally predicted variances.
We demonstrate an initial study showing our model's ability to quantify its own uncertainty and switch to full-beam sweeps in Appendix~\ref{app:uncertaintyOracle}. While these results already show promise, we leave a more thorough analysis of how to strategically switch between different beam-selection algorithms for future work.

\begin{table}[t!]
    \caption{Ablation across codebook sizes and basestation position. 
    Reported are the percentage of the maximum cumulative RSS / raw cumulative RSS.
    Test results shown here on unseen trajectories.}
    \label{tab:size_position_ablation}
    \centering
    \resizebox{\linewidth}{!}{
    \begin{tabular}{ccccccc}\toprule
                      &\multicolumn{3}{c}{low attenuation}&\multicolumn{3}{c}{high attenuation}\\\midrule
                      &$5\times 5$ & $9\times 9$ &  $11\times 11$ &$5\times 5$ & $9\times 9$ &  $11\times 11$ \\\midrule
         BS1& 27\% / 0.63     &  13\% / 1.16     &  28\%  / 6.03       & 28\% / 0.61     &  15\% / 1.01      &   50\% / 10.46      \\
         BS2& 92\% / 7.91     &  91\% / 27.90     &  81\% / 65.13        & 94\% / 7.36     &  92\% / 28.18    &   70\% / 48.82       \\
         BS3& 90\% / 4.67     &  82\% / 13.96     &  91\% / 42.61        & 94\% / 4.73     &  92\% / 16.27     &   91\% / 43.32       \\
         BS4& 93\% / 4.43     &  68\% / 10.27     &  57\% / 18.11        & 93\% / 4.38     &  69\% / 9.29     &   55\% / 17.51        \\\bottomrule
    \end{tabular}
    }
\end{table}

\subsection{Impact of Codebook Size and Basestation Position}
\label{sec:Impact_size_env}

We consider different configurations of the \emph{beam and basestation} topology. Specifically we study another ``blocker ref'' scenario, but now choose from 3 different codebook sizes (5$\times$5, 9$\times$9, 11$\times$11) and 4 different basestation positions (BS1, BS2, BS3, BS4). We ablate the effect of blocker strength by considering both a low attenuation 
and a high attenuation blocker (infinite strength). Finally, we increase the trajectory length to 390 steps to validate long-range inference performance.
Hence, in total we train 3$\times$4$\times$2=24 policies. This experiment design demonstrates that our approach works on a variety of settings with different underlying transition distributions and codebook sizes.

Table~\ref{tab:size_position_ablation} shows the test results on unseen trajectories.
This benchmark has roughly twice the number of timesteps per trajectory than the previous environments and considers two variants across three different codebook sizes.
We found that we needed to increase the amount of optimization steps to account for the increase in the number of beams:
Specifically, we increased the amount of compute used by $2\times$, while keeping the amount of environment interactions the same.

We additionally report top-3 accuracy results in Table~\ref{tab:size_position_top_3}.
We report a top-3 accuracy due to the high number of beam selections with nearly identical performance. We found that top-3 accuracy generally correlates well with empirical RSS distributions on our simulation.

We see high performance in all configurations apart from BS1, which is placed such that it points directly into the blocker. This means that the difference between an optimal and a suboptimal beam is small as even the optimal beam can barely see the UE.
As no beam clearly dominates another, the model continuously explores different beams rather than sticking to one slightly better beam.
We also observe a lower accuracy in high dimensional codebooks, which is however balanced out by the \emph{much} higher possible transmission rates possible for larger codebooks.
This is expected as more available actions make the selection problem harder, but the optimal beam has higher throughput.
The normalized RSS drops for larger codebooks, while the actual received power increases substantially.
\begin{table}[t!]
    \caption{Top-3 accuracy across codebook sizes and basestation position. 
    }
    \label{tab:size_position_top_3}
    \centering
    \resizebox{\linewidth}{!}{
    \begin{tabular}{ccccccc}\toprule
                      &\multicolumn{3}{c}{low attenuation}&\multicolumn{3}{c}{high attenuation}\\\midrule
                      &$5\times 5$ & $9\times 9$ &  $11\times 11$ &$5\times 5$ & $9\times 9$ &  $11\times 11$ \\\midrule
         Basestation 1& 30\%     &  16\%      &  36\%        & 30\%     &  16\%      &   56\%       \\
         Basestation 2& 98\%    &  95\%     &  88\%        & 96\%     &  95\%     &   78\%       \\
         Basestation 3& 95\%     &  89\%     &  96\%        & 97\%     &  96\%     &   97\%       \\
         Basestation 4& 96\%     &  75\%     &  63\%        & 96\%     &  71\%     &   56\%        \\\bottomrule
    \end{tabular}
    }
\end{table}
\begin{table}[b!]
    \centering
    \caption{Expected cumulative RSS as a percentage of maximally achievable over time. We train and evaluate models on different speeds and observe consistent performance across all different speed ranges.}
    \label{tab:speedResults}
    \begin{tabular}{lrrrr}
    \toprule
     & 1.11\si{m/s} & 2.22\si{m/s} & 3.33\si{m/s} & 4.44\si{m/s} \\
    \midrule
    model@1.11\si{m/s} & 74.23\% & 73.27\% & 75.12\% & 75.65\% \\
    model@2.22\si{m/s} & 74.84\% & 73.99\% & 74.91\% & 74.57\% \\
    model@3.33\si{m/s} & 75.04\% & 74.25\% & 75.22\% & 74.87\% \\
    model@4.44\si{m/s} & 74.48\% & 74.10\% & 74.81\% & 75.18\% \\
    \bottomrule
    \end{tabular}
\end{table}

\subsection{Impact of movement speed}\label{sec:movementspeedBench}

We also evaluate the impact of UEs moving at drastically different speeds. 
We generate a set of trajectories and train on different sampling rates (to account for different movement speeds of the UE):
Sampling every timestep of the trajectories leads to a ``base speed'' (slowest), while skipping every N entries gives an effective speed of $N\times\text{base speed}$.
We set our base speed at $1.11\si{m/s}$ and compare $N=4$ speed levels each with a $5\times 5$ codebook.
To adequately test this, we generate long trajectories with 500 timesteps, such that ``skipping'' steps for faster moving UEs still have enough timesteps to allow for meaningful comparison.
Our models were first trained on random trajectories in the environment with a total training time of on average $2.5\si{h}$.\footnote{Training time depends on the speed since faster trajectories have fewer timesteps to train over} After training, inference time is negligible.
Notice that since we only record the RSS values for each beam, doppler shifts do not affect the measurements.

The results are shown in Table~\ref{tab:speedResults}. Our method is robust against changes in UE speed both during training and inference. 
We want to note that the absolute velocities are not too important for this analysis: only the relative change between speeds is important from a selection point of view, since our method makes no assumption on the physical or antenna configuration.
The main change for faster trajectories is that faster trajectories lead to faster changes in the codebook values, but the relative speedup in codebook changes only corresponds to how much of a multiple of the training speed our method is evaluated at.


\subsection{Impact of Goal-model size}\label{sec:GoalModelAblation}

\begin{figure}[t!]
    \centering
    \begin{tikzpicture}
    \begin{axis}[xlabel={exploration steps},ylabel={expected total RSS per episode}, ylabel near ticks,ytick distance=1,
    xlabel near ticks,width=0.85\linewidth,legend style={at={(0,1)},anchor=north west}]
    \addplot table [x={eval step}, y={5 goal depth - eval reward}, col sep=comma, smooth,x expr=\thisrowno{0}*2500] {run-csvs/goalModelAblation.csv};
    \addplot table [x={eval step}, y={4 goal depth - eval reward}, col sep=comma, smooth,x expr=\thisrowno{0}*2500] {run-csvs/goalModelAblation.csv};
    \addplot table [x={eval step}, y={3 goal depth - eval reward}, col sep=comma, smooth,x expr=\thisrowno{0}*2500] {run-csvs/goalModelAblation.csv};
    \addplot table [x={eval step}, y={2 goal depth - eval reward}, col sep=comma, smooth,x expr=\thisrowno{0}*2500] {run-csvs/goalModelAblation.csv};
    \legend{depth 5,depth 4,depth 3,depth 2};
    \end{axis}
    \end{tikzpicture}
    \caption{Performance comparison between different goal model sizes on the ``no house'' environment. We observe significant performance increases up to a depth of 4.}
    \label{fig:GoalModelAblation}
\end{figure}
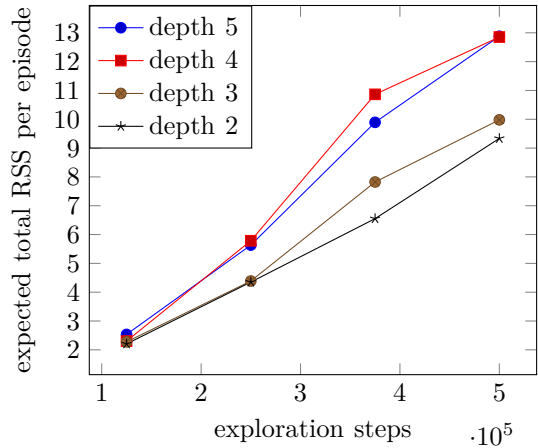

One of the most important elements of the search model we propose is the implicit correction towards domain shift introduced by the goal model.
For this reason, we compare different numbers of ``goal state'' layers (top of \cref{fig:intro} against each other.

As shown in Fig~\ref{fig:GoalModelAblation}, performance improves up until a depth of 4 GRU layers, after which the improvement stops.
We hypothesize that this is due to the model getting close to representing the ``true'' goal belief distribution.
Exceeding the capacity needed to approximate the goal distribution will not improve overall performance as even in the limit of perfect belief-state representation, the model would not be able to accurately capture the true goal due to partial observability.

\subsection{Comparison against the State of the Art}\label{sec:SOTA_comparison}

We compare our method with two contemporary state-of-the-art methods in intelligent beam selection and RL respectively.
\textbf{Beam selection.} We compare against an online Bayesian optimization scheme proposed by \citep{Maggi2023TrackingTB}, which iteratively samples beams from the codebook according to an acquisition function over a Gaussian Process (GP) fit on the measured RSS.
In contrast to our work, \citet{Maggi2023TrackingTB} must sample multiple beams per timestep (configured via an ``overhead'' function).
As we only samples one beam per timestep\footnote{Recent work by \citet{Karbasi2021ParallelizingTS} shows that Thompson sampling in the ordinary MAB setting can also be parallelized. We believe an analogous scheme also works for RMABs but leave the characterization of this scheme for future work.}, we fix the number of beams tried in \citet{Maggi2023TrackingTB} and compare different per-timestep sampling budgets. For this we take the same test set used in \Cref{sec:movementspeedBench}.

\begin{table}[t!]
    \caption{GP baseline~\citep{Maggi2023TrackingTB} for different speeds and different number of sampled beams per timestep. Drawing more than 4 samples per timestep is intractable due to exploding computational and memory complexity in the GP. For multiple samples we always assume the \emph{best} beam is used for transmission.}
    \centering
    \begin{tabular}{lrrrr}
    \toprule
     & 1.11\si{m/s} & 2.22\si{m/s} & 3.33\si{m/s} & 4.44\si{m/s} \\
    \midrule
    samples 1 & 65.85\% & 65.68\% & 67.51\% & 66.99\% \\
    samples 2 & 78.30\% & 77.42\% & 78.84\% & 78.04\% \\
    samples 3 & 81.44\% & 79.07\% & 79.83\% & 79.35\% \\
    samples 4 & 88.18\% & 86.27\% & 87.16\% & 85.80\% \\
    \bottomrule
    \end{tabular}
    \label{tab:Baseline}
\end{table}
The results are shown in Table~\ref{tab:Baseline}. Our method is consistently better than the GP based baseline in the 1 sample case.
However, once multiple samples per timestep can be drawn, the GP baseline starts closing in and exceeding our method, though at the expected increase in real-world sampling cost: Trying 2 different beams corresponds to sampling close to $10\%$ of the codebook for a $5\times 5$ antenna array.

One core issue with GP based solutions is the huge computational burden involved in fitting GPs to data: Computing the weights of a GP for a dataset with $N$ elements corresponds to fitting kernel hyperparameters by constructing and then inverting a large $N\times N$ matrix multiple times.
This quickly becomes intractable in both memory and compute, especially on edge devices with real time guarantees.
For this method, the dataset grows with $\text{num\_timesteps}\times\text{samples\_per\_timesteps}$. This means the ``samples 4'' case would yield a matrix of size $2000\times 2000$ to construct and invert multiple times.

In practical terms, this means that computing the results of Table~\ref{tab:Baseline} on an Intel Xeon Gold 5120 server-CPU across 25 parallel threads takes approx. 13 hours, while inference for our method in Table~\ref{tab:speedResults} across \emph{a single thread} takes $\approx 20 \si{min}$ on a consumer CPU.
This means our method is approximately $975\times$ more efficient to infer, with that advantage \emph{increasing} with larger time horizons.
Further, we expect our runtime to be improveable by e.g. applying pruning or quantization techniques to the model.

From a complexity point of view, inferencing our method scales in time with order $O(1)$ while GP solutions are constrained to scaling in $O(T^3)$.
This means that our method is realtime capable, while GPs generally are not.\footnote{There are ways of speeding up Gaussian processes via, for instance, Nyström approximation, or using sparse variational GPs~\citep{pmlr-v5-titsias09a}. This usually comes with a drop in performance. General scaling of GPs to big data remains a major challenge, especially in online learning scenarios.}

\begin{table}[t!]
    \caption{Evaluation from the high-attenuation set for BS4 seen in Table~\ref{tab:size_position_ablation}. 
    The GP method has significantly lower performance when compared to ours while having higher computational and memory complexity}
    \label{tab:GPSizeAblation}
    \centering
    \begin{tabular}{ccccc}\toprule
              & 1 sample & 2 samples & 3 samples & 4 samples\\\midrule
        5x5   & 10.04\%  & 18.27\%   & 23.61\%   & 28.97\%  \\
        9x9   & 4.93\%   & 9.70\%    & 12.09\%   & 14.06\%  \\ 
        11x11 & 2.07\%   & 5.09\%    & 5.69\%    & 6.88\%   \\\bottomrule
    \end{tabular}
\end{table}

However, an even bigger concern for the GP based solution is how well these methods scale to larger codebook sizes.
GP based search is well known to behave poorly when increasing the number of dimensions to search over.
We evaluate \citet{Maggi2023TrackingTB} on the existing $5\times 5$, $9\times 9$ and  $11\times 11$ codebook grids used for the high attenuation tests in Table~\ref{tab:size_position_ablation}.
As can be seen in Table~\ref{tab:GPSizeAblation}, the GP based search behaves poorly compared to ours, which we believe to be due to many irrelevant beams existing in the set\footnote{Note that the $5\times 5$ results are worse than in the ``moving UEs case'' due to fewer beams with high performance}.
While our search can exploit prior knowledge in how beamsets correlate and, based on that information, discard low value choices, \citet{Maggi2023TrackingTB} has to discover these correlations at inference time, severely increasing the number of samples needed for good coverage.
To get good results on large codebooks one needs to \emph{significantly} increase the number of beam-measurements per timestep. This would increase the memory and compute costs dramatically.
This also implies an implicit dependence between codebook size and time/memory complexity.


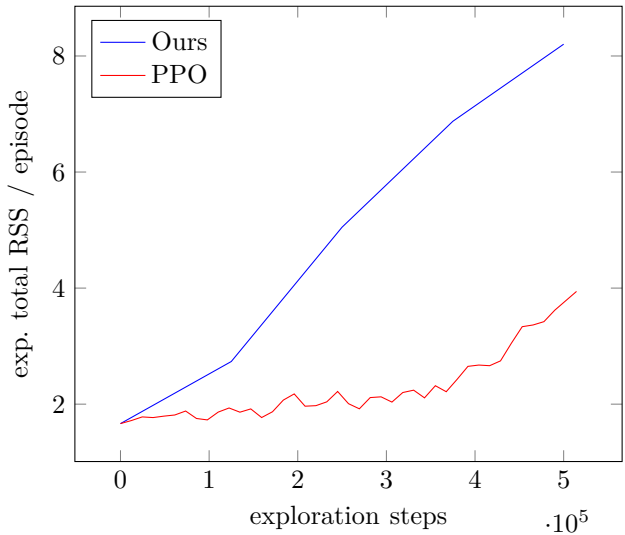
\begin{figure}[!htb]
    \vspace{-1mm}
    \begin{tikzpicture}
    \begin{axis}[xlabel={exploration steps},ylabel={exp. total RSS / episode}, ylabel near ticks,
    xlabel near ticks,width=\linewidth,legend pos=north west]
    \addplot[blue] table [x={Eval-step}, y={reward}, col sep=comma,x expr=\thisrowno{0}*2500] {run-csvs/noisy-with-house_zeros.csv};
    \addplot[red] table [x={Step},x expr=\thisrowno{0}*4.9*2500, y={ep_rew_mean}, col sep=comma, mark=none] {run-csvs/PPO-house.csv};
    \legend{Ours,PPO}
    \end{axis}
    \end{tikzpicture}  
    \caption{Total RSS of PPO vs. our approach.}
    \label{fig:house_result_ppo}
\end{figure}

\textbf{Reinforcement Learning.} On the RL side, we further demonstrate our method's high performance by comparing to a tuned PPO agent~\citep{Schulman2017ProximalPO} with the implementation from~\citep{stable-baselines3} against our approach.
PPO is a state-of-the-art RL method consisting of a policy gradient that is constrained to stay within a KL-Ball to ensure policy updates do not put undue confidence on low amounts of data (see \citet{Schulman2015TrustRP,Schulman2017ProximalPO}).
Both our agent and the tuned PPO agent are given a training budget of $500,000$ steps on the ``blocker'' environment.
Fig.~\ref{fig:house_result_ppo} shows the achieved RSS values on a validation set over the course of training, with the x-axis being the total number of exploration steps seen by each algorithm.
Our method outperforms the PPO-baseline right from the beginning, reaching PPO's final performance after having seen only a little more than half of the total exploration budget.
Under equal exploration budgets for our method and PPO, we reach almost double the baseline's RSS.

\section{Conclusion}\label{section:conclusion}

We proposed a novel method for solving challenging POMDPs where a critical challenge is to trade-off belief-state estimation and reward maximization.
We solve these problems by recursively deploying a neural network as an approximate posterior into an environment, running online restless multi-armed bandit inference, and using the gathered information to update the network weights for the next rollout.
We prove the efficiency of our approach by evaluating our method in hard beam-tracking problems across many different environment and network configurations.
Our method has the potential to improve on existing methods for online RMAB inference in general.
\bibliographystyle{IEEEtranN}
\bibliography{bibliography}


\begin{figure*}
    \centering
    \begin{subfigure}[t]{0.45\linewidth}
    \begin{tikzpicture}
        \begin{axis}[
            xlabel={timesteps},
            ylabel={expected uncertainty},
            title={Uncertainty of BS4 for size $11\times11$},
            width=\textwidth
        ]
        \addplot[blue, mark=none] table[x=timestep, y=lowatt,col sep=comma] {{beamUncertainty/beamUncertainty.csv}};
        \addplot[red, mark=none] table[x=timestep, y=highatt,col sep=comma] {{beamUncertainty/beamUncertainty.csv}};
        \legend{low attenuation, high attenuation}
        \end{axis}
    \end{tikzpicture}
    \end{subfigure}%
    \begin{subfigure}[t]{0.45\linewidth}
    \centering
    \begin{tikzpicture}
        \begin{axis}[
            xlabel={timesteps},
            title={Uncertainty of BS2},
            cycle list name= color list,
            width=\textwidth
        ]
        \addplot+[mark=none] table[x=timestep, y=5x5_high,col sep=comma] {{beamUncertainty/beam_uncertainty_BS2.csv}};
        \addplot+[mark=none] table[x=timestep, y=5x5_low,col sep=comma] {{beamUncertainty/beam_uncertainty_BS2.csv}};
        \addplot+[mark=none] table[x=timestep, y=9x9_high,col sep=comma] {{beamUncertainty/beam_uncertainty_BS2.csv}};
        \addplot+[mark=none] table[x=timestep, y=9x9_low,col sep=comma] {{beamUncertainty/beam_uncertainty_BS2.csv}};
        \addplot+[mark=none] table[x=timestep, y=11x11_low,col sep=comma] {{beamUncertainty/beam_uncertainty_BS2.csv}};
        \addplot+[mark=none] table[x=timestep, y=11x11_high,col sep=comma] {{beamUncertainty/beam_uncertainty_BS2.csv}};
        \legend{5x5 high, 5x5 low, 9x9 high, 9x9 low, 11x11 low, 11x11 high}
        \end{axis}
    \end{tikzpicture}
    \end{subfigure}
    \caption{Uncertainty of beam quality over time as measured using our XGBoost model.}
    \label{fig:uncertainty_graph}
\end{figure*}
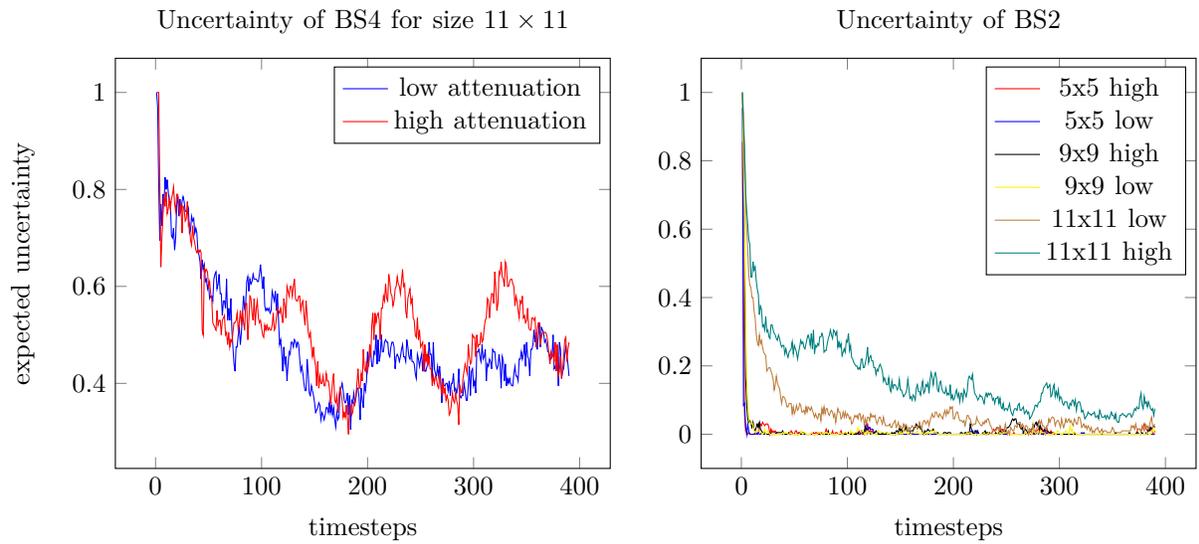
\newpage
\appendix

\subsection{Uncertainty based Oracle evaluation }\label{app:uncertaintyOracle}

One of the great benefits of bayesian systems is their ability to predict their own uncertainty for any prediction.
In the context of beam selection one can exploit this property by selecting a different algorithm once our method self-reports high uncertainty.
To provide initial results on this design, we train an XGBoost~\citep{Chen2016XGBoostAS} model that predicts the likelihood of being outside the top-5 arms, \emph{solely from the amplitude of the latent noise}.
To set the hyperparameters, we use hyperopt~\citep{pmlr-v28-bergstra13} on 5-fold cross-validation over the existing training trajectories to select for a maximal f1-score.

Using this XGBoost model, we analyze our meta-bandit's ability to measure its own uncertainty on beam choice quality. We accomplish this by training a classifier to predict whether the chosen arm is inside the top-5 best arms, based on the variance of the latent distribution
The results are shown in Fig.~\ref{fig:uncertainty_graph}.
One can clearly see that the model has high uncertainty at the beginning of exploration, and quickly reduces due to online learning.
This uncertainty management allows for automatic failover to an \eg exhaustive beam-sweep in case the predicted model's uncertainty rises above a threshold.
To maximize performance, one would have to train our model with this ``escape hatch'' already in mind, as calling an external oracle leads to a distribution shift in our timeseries estimate of the goal and arm posteriors.
We leave a thorough study of this for future work.

\begin{table}[t!]
    \centering
    \caption{Oracle based evaluation on Basestation 4 results}
    \resizebox{\linewidth}{!}{
    \begin{tabular}{ccccccc}\toprule
                              &\multicolumn{3}{c}{low attenuation}&\multicolumn{3}{c}{high attenuation}\\\midrule
                              &$5\times 5$ & $9\times 9$ &  $11\times 11$ &$5\times 5$ & $9\times 9$ &  $11\times 11$ \\\midrule
    Percentage of total RSS                      &90.47\%      &  66.88\%     &  74.11\%        &90.73\%      & 68.81\%      &  76.36\%        \\
    Percentage Call-to-Oracle & 1\%        &  5\%       &  49\%          & 1\%        & 4\%        &  53\%          \\\bottomrule
    \end{tabular}
    }
    \label{tab:oracle_bs4}
\end{table}

Inspecting our results in Table~\ref{tab:oracle_bs4}, we observe that both $5\times 5$ and $9\times 9$ only rarely calls into the oracle and overall looses a tiny bit of performance when compared to the baseline without an oracle.
We hypothesize that this is specifically due to the distribution shift and the low amount of information gained from the oracle: Both of these models are already very close to the theoretical maximum and the information gain in these small beam-id grids is relatively minimal. 
Calling into the oracle is most likely just confusing the model due to the distribution shift.
In the $11\times 11$ case, we see significant performance improvements from having an oracle backup.
This makes sense since getting the ``correct'' beam yields much more information than in the smaller grids, and the $11\times 11$ configuration is further away from the true optimum.



In general, we argue that such an uncertainty based prediction model would have to be introduced into the model training process, for example by adding in an additional selection option that provides low reward, but calls the oracle for increased long-term performance.
However, as can be seen in our $11\times 11$ model, even naive implementations of uncertainty quantification can greatly boost the final performance.
We leave investigations of more efficient versions of this approach for future works.

\end{document}